\documentclass[11pt]{article}

\usepackage[final]{acl}

\usepackage{times}
\usepackage{latexsym}
\usepackage[T1]{fontenc}
\usepackage[utf8]{inputenc}
\usepackage{microtype}
\usepackage{inconsolata}
\usepackage{graphicx}
\usepackage{booktabs}
\usepackage{multirow}
\usepackage{amsmath}
\usepackage{amssymb}
\usepackage{array}
\usepackage{xspace}
\usepackage[normalem]{ulem}

\usepackage{xcolor}
\definecolor{RIblue}{RGB}{0,102,204}
\definecolor{RLbrown}{RGB}{150,75,0}

\title{What Do Biomedical NER and Entity Linking Benchmarks Measure? A Corpus-Centric Diagnostic Framework}

\author{Robert Leaman\thanks{\, These authors contributed equally} \\
   \\
  \texttt{Robert.Leaman@nih.gov} \\\And
  Rezarta Islamaj\footnotemark[1] \\
  National Library of Medicine, Bethesda, MD \\
  \texttt{Rezarta.Islamaj@nih.gov} \\\And
  Zhiyong Lu \\
   \\
  \texttt{Zhiyong.Lu@nih.gov} \\
  }

\begin{document}
\maketitle

\begin{abstract}
Biomedical named entity recognition (NER) and entity linking (EL) strongly depend on annotated corpora, but the utility of these resources for benchmarking is often assumed rather than 
characterized.
We present a corpus-centric framework for diagnosing benchmark-relevant properties directly from corpus annotations, concept links, train-test splits, document metadata, and terminology mappings. The framework organizes standardized statistics into five families: (1) scale, density and label distribution, (2) lexical and conceptual structure, (3) train-test overlap, (4) metadata composition, and (5) terminology coverage where applicable.
Applying the framework to nine corpora spanning diseases, chemicals, and cell types, we find that corpus properties can differ substantially, even when they address the same apparent task.  
We find differences in the evaluation signal they provide, the generalization demands they impose, the degree of train–test reuse they permit, and the regions of biomedical literature and concept space they represent.
These differences suggest that commonly reported corpus statistics can be insufficient to characterize what biomedical NER and EL benchmarks evaluate.
We argue that corpus-centric diagnostics provide a practical framework for analyzing corpora beyond surface descriptors such as corpus size and entity type, for identifying potential transfer risks, and for interpreting the scope of benchmarking conclusions.
We release the framework as open-source code\footnote{https://github.com/NLM-DIR/CorpusBenchmarking} with an interactive dashboard to support reproducing our analyses and characterizing additional corpora.
\end{abstract}

\section{Introduction}

Extracting structured information from biomedical literature requires systems to identify and link entities (e.g., genes, diseases, chemicals) to standardized identifiers. These grounding tasks---named entity recognition (NER) and entity linking (EL)---remain critical in the large language models (LLM) era to ensure outputs are auditable, comparable, and reusable.

Manually annotated corpora serve both as training data and as evaluation benchmarks~\citep{kim2003,collier2004,morgan2008,lu2011,wei2013,dogan2014,krallinger2015,li2016,islamaj2021chem,islamaj2021gene,bada2012,herrero-zazo2013,wei2016,miranda2023}. When used as benchmarks, they function as measurement instruments: the key question is not only whether annotations are correct, but what capabilities the benchmark tests and whether conclusions transfer to intended use cases.

This distinction matters because benchmark utility is task- and domain-dependent. A corpus can be carefully annotated yet too narrow, homogeneous, or leaky across splits to support informative evaluation. Biomedical natural language processing (NLP) values rare, specialized, and emerging concepts, making evaluation sensitive to which entities, subdomains, time periods, and document types are represented. Without characterizing the \emph{corpus domain} and the target \emph{application domain}, it is difficult to distinguish generalization issues from corpus-specific artifacts, leakage, or domain mismatch. Benchmark corpora have primarily been compared by size, entity type, or reported system performance. These descriptors do not reflect overlap risks, lexical difficulty, domain bias, or concept coverage. 

\begin{figure*}[t]
  \centering
  \includegraphics[width=\linewidth]{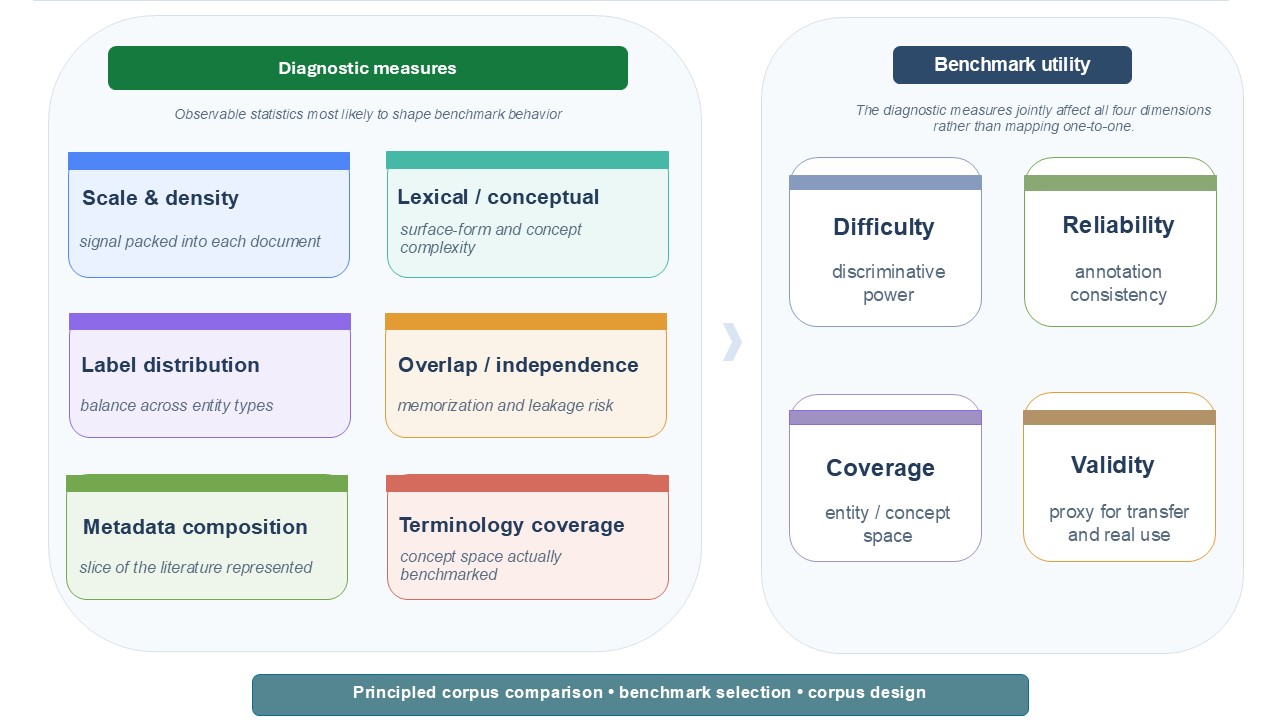}
  \caption{
  Corpus diagnostic framework. Entity-annotated corpora are converted into a common representation, enabling computation of statistics over annotations, identifiers, and metadata. These statistics characterize scale and density, lexical and conceptual variation, train-test overlap, metadata composition, and terminology coverage, enabling principled comparison of biomedical NER and EL benchmarks. 
  }
  \label{fig:framework}
\end{figure*}

To address this gap, we introduce a corpus-centric framework that computes standardized statistics directly from annotations, concept links, corpus splits, metadata, and terminologies (Figure~\ref{fig:framework}).
These statistics provide a multidimensional analysis for PubMed- and PMC-based NER and EL corpora: density indicates how much evaluation signal is available; lexical and conceptual variation indicate what kinds of generalization are required; overlap reveals leakage risk; metadata composition characterizes the literature represented; and terminology coverage indicates which parts of the concept space are represented.

Applied to nine corpora spanning diseases, chemicals, and cell types, the framework shows that resources similar by task label or size can differ substantially across these signals. Our contributions are: (1) a corpus-as-instrument framing for biomedical NER and EL benchmarks; (2) a practical framework of corpus-centric diagnostics; (3) an analysis showing how structural differences inform evaluation sensitivity, leakage risk, coverage, and transferability; and (4) open-source code and an interactive dashboard for reproducing our results and for analyzing new corpora.

\section{Related Work}

Biomedical entity annotation corpora have proliferated over the years, yet they are routinely treated as benchmarks without any systematic analysis of what they actually measure. Early efforts, such as GENIA ~\citep{kim2003} established large-scale manual mention annotation with fine-grained semantic categories, later simplified for shared tasks such as JNLPBA~\citep{collier2004}. Subsequent corpora introduced entity normalization, typically targeting a single entity type: diseases (NCBI Disease), chemicals (CHEMDNER, BC5CDR), genomic variants (tmVar) and genes (BioCreative challenges)~\citep{dogan2014,krallinger2015,li2016,wei2013,morgan2008}. Resources such as CRAFT~\citep{bada2012} broadened the scope to multi-entity, full-text annotation. More recent corpora---including NLM-Chem, BioRED, and CellLink--have further expanded document and entity coverage, and incorporate richer annotation structures such as relations~\citep{islamaj2021chem,islamaj2024b,rotenberg2026}. Despite being widely used together, these corpora vary considerably in document type, annotation density, normalization support, temporal range, and domain focus---differences that are rarely examined in terms of what each benchmark actually evaluates.

NLP research has shown that dataset properties can distort benchmark interpretation. Work on saturation motivates aggregated benchmarks such as GLUE and SuperGLUE~\citep{wang2018,wang2019}; work on artifacts, leakage, and memorization shows that apparent improvements can reflect shortcuts or overlap rather than the intended capability~\citep{gururangan2018,liang2023,Tutubalina2020}; and HELM emphasizes multi-metric evaluation~\citep{liang2023}. Biomedical suites such as BLUE, BLURB, and BigBIO standardize cross-task evaluation~\citep{Peng2019, Gu2021,fries2022}, but generally treat corpora as fixed inputs rather than explaining how corpus properties shape validity or transferability.

Annotation quality measures, particularly inter-annotator agreement (IAA), assess consistency but not benchmark scope. Prior work distinguishes agreement across span boundaries, labels, and concept links~\citep{artstein2008}, recommends F1 for span-based tasks lacking a well-defined negative class~\citep{hripcsak2005}, and notes that disagreement may reflect ambiguity, error, or guideline limitations~\citep{aroyo2015,uma2021}. High agreement is necessary but not sufficient: simplifying annotation can increase agreement while removing realistic ambiguity~\citep{hovy2010}. Corpus papers often report counts and distributions, but these statistics are rarely organized around evaluation claims. Our framework connects such descriptions to overlap, memorization, domain shift, annotation scope, and terminology coverage.

\section{Methods}

\subsection{Framework Overview and Representation}

Our framework characterizes NER and EL corpora through four stages: conversion to a shared representation, filtering, metric computation, and visualization. Corpora are standardized into documents containing text, optional metadata, and annotations (spans, surface forms, labels, and linked concept identifiers).

This design supports both NER-only and NER+EL PubMed- or PMC-based corpora. NER-only corpora are evaluated on text-, span-, and mention-level statistics, while EL-supported datasets additionally yield concept-level diagnostics. Metrics are computed over configurable corpus bundles, comparison suites, entity scopes, and train/dev/test partitions to enable interpretable comparisons across heterogeneous datasets.

\subsection{Corpora Analyzed}

We apply the framework to nine corpora spanning diverse entity types (e.g., diseases, chemicals, cell types) and document scopes (abstracts, captions, full-text): AnatEM~\citep{pyysalo2014}, BC5CDR~\citep{li2016}, BioID~\citep{arighi2017}, CHEMDNER~\citep{krallinger2015}, CRAFT~\citep{bada2012}, CellLink~\citep{rotenberg2026}, JNLPBA~\citep{collier2004}, NCBI-Disease~\citep{dogan2014}, and NLM-Chem~\citep{islamaj2021chem}. Where public test data were unavailable or original annotation layers were altered, we utilized the closest documented subset and note these limitations alongside the relevant results.

\subsection{Diagnostic Statistics}

The framework computes corpus statistics across five families to diagnose benchmark properties prior to system evaluation:

\begin{itemize}
\item \textbf{Scale, Density, and Label Distribution}: We compute total documents, tokens, annotations, and unique mentions/identifiers per document.
\item \textbf{Lexical and Conceptual Structure}: For normalized corpora, we measure mention ambiguity (the number of distinct label/link pairs mapped to a single surface form) and identifier variation (the number of distinct surface forms mapped to a label/link pair). These distinguish a benchmark's demand for contextual disambiguation versus its demand for recognizing diverse lexical realizations.
\item \textbf{Train-Test Overlap}: To assess leakage and memorization risk, we compute Jaccard overlap between train and test splits at four abstraction levels: general token vocabulary, tokens inside entity mentions, exact mention strings, and concept identifiers.
\item \textbf{Metadata Composition}: We profile the represented literature slice via temporal statistics (publication year ranges and distributions) and journal diversity (unique journals and top-journal concentration).
We derive broad topic profiles from article 
Medical Subject Headings (MeSH)~\citep{Lipscomb2000} topics where available, falling back to NLM Catalog MeSH journal topics if necessary.
\item \textbf{Terminology-Aware Coverage}: For diseases and chemicals, we link identifiers to their respective concepts in MeSH, MONDO~\citep{Vasilevsky2026}, or ChEBI~\citep{Malik2026}; for cell types, we support Cell Ontology (CL)~\citep{Tan2026}. 
We quantify vocabulary coverage by analyzing the distribution of annotations across high-level branches and compute hierarchy depth as a proxy for concept specificity.
\end{itemize}

\subsection{Implementation}

The framework is implemented as an open-source, YAML-configurable Python pipeline that outputs structured JSON statistics. It includes acquisition specifications for downloading, extracting, converting, caching, and validating expected corpus files. Input support includes 
BioC XML~\citep{Comeau2013,Comeau2019}, PubTator~\citep{wei2024}, BRAT/standoff~\citep{Stenetorp2012}, and Knowtator~\citep{Ogren2006} annotations, with registry-based extension points for additional loaders and metrics. Ontologies in OBO  format are directly supported as terminology sources. The accompanying self-contained HTML/JavaScript dashboard\footnote{https://nlm-dir.github.io/CorpusBench-marking/dashboard.html} combines scale, overlap, metadata, terminology, and entity-scope views to reproduce our analyses and evaluate new corpora.

\section{Results}

We use the nine corpora to illustrate how corpus-centric diagnostics clarify the evaluation role of a benchmark. The goal is not to rank corpora, but to demonstrate that datasets with similar task labels often function as different measurement instruments: they expose systems to different volumes of evaluation signal, different forms of lexical and conceptual generalization, different leakage risks, and different regions of the biomedical literature and concept space.

\subsection{Scale, annotation density, lexical and conceptual variation}

Table~\ref{tab:basic} reports statistics for nine heterogeneous biomedical corpora.
These measurements show that corpora differ fundamentally in the structural nature of the evaluation signal they provide. 
Annotation density varies widely across corpora, reflecting differences in document scope, annotation unit, and entity scope. Dense full-text corpora such as NLM-Chem and CRAFT provide many labeled decisions per article, whereas passage-, abstract-, and caption-based corpora distribute fewer decisions across more sampled text units. This raw density is useful because it affects the number of system decisions contributing to an evaluation estimate. However, annotations from the same full-text article are not necessarily independent: repeated mentions, recurring identifiers, and section-specific language can increase decision volume without proportionally increasing lexical, conceptual, or contextual diversity. Density should therefore be interpreted as signal concentration, not as a direct measure of task difficulty or benchmark quality. In this sense, full-text corpora evaluate behavior over long-document contexts and repeated real-world usage, while shorter-unit corpora may provide broader sampling of independent contexts per annotation. Concept-level diagnostics further define what the instrument is calibrated to measure. Variation ranges from 1.48 surface forms per label/link pair in BioID to 3.74 in CellLink, indicating differing demands on lexical generalization.

\begin{table*}[t]
\centering
\small
\setlength{\tabcolsep}{4pt}
\begin{tabular}{lp{1.4cm}rrrrrrrp{1.3cm}rrr}
\toprule
\textbf{Corpus} & \textbf{Doc. type} & \textbf{Docs} & \textbf{Tokens} & \textbf{E} & \textbf{Total} & \textbf{Ann/} & \textbf{Men./} & \textbf{IDs/} & \textbf{ID} & \textbf{Ambi-} & \textbf{Vari-} \\
& & & & & \textbf{Ann.} & \textbf{doc} & \textbf{doc} & \textbf{doc} & \textbf{vocab} & \textbf{guity}\textsuperscript{a} & \textbf{ation}\textsuperscript{b} \\
\midrule
AnatEM    & abstracts & 1,212  & 259,510   & 12 & 13,701  & 11.3   & 7.2   & ---   & ---           & ---   & ---  \\
BC5CDR    & abstracts & 1,500  & 297,019   & 2  & 29,271  & 19.5   & 9.4   & 6.9   & MeSH          & 1.02 & 2.47 \\
BioID     & captions & 13,697 & 771,248   & 8  & 102,742 & 7.5    & 4.9   & 5.2   & mixed\textsuperscript{c} & 1.35 & 1.48 \\
CHEMDNER  & abstracts & 10,000 & 2,092,491 & 1  & 84,331  & 8.4    & 4.6   & ---   & ---           & ---   & ---  \\
CRAFT     & full text & 97     & 652,168   & 11 & 99,623  & 1027.0 & 241.9 & 149.4 & mixed\textsuperscript{c} & 1.02 & 2.33 \\
CellLink  & passages & 2,003  & 227,490   & 3  & 14,731  & 7.3    & 6.0   & 4.1   & CL            & 1.07 & 3.74 \\
JNLPBA    & abstracts & 2,404  & 564,660   & 5  & 59,963 & 24.9  & 16.5 & ---  & ---           & ---  & ---    \\
NCBI-Disease & abstracts & 793 & 169,561  & 1  & 6,892  & 8.7   & 5.1  & 3.2  & MeSH/OMIM     & 1.01 & 2.75 \\
NLM-Chem  & full text & 150    & 789,532   & 1  & 38,339 & 255.6 & 54.3 & 34.1 & MeSH          & 1.02 & 2.42 \\
\bottomrule
\end{tabular}
\vspace{4pt}
\par\noindent\footnotesize

\textsuperscript{a}~\textbf{Ambiguity}: mean label/link pairs per unique mention string. Values near 1.0 indicate most mentions map to a single label/link pair.
\textsuperscript{b}~\textbf{Variation}: mean surface forms per label/link pair. Reported only for corpora with concept-level identifiers; AnatEM, CHEMDNER, and JNLPBA are excluded (n/a).
\textsuperscript{c}~BioID and CRAFT link entity types to many ontology identifier resources.
\caption{Basic statistics for each corpus. E = number of annotated entity types; Men./doc = unique mention strings per document; IDs/doc = unique concept identifiers per document. CellLink values reflect the currently released annotated train and development partitions.}
\label{tab:basic}
\end{table*}

\subsection{Train-Test Overlap}

\begin{figure*}[t]
  \centering
  \includegraphics[width=\linewidth]{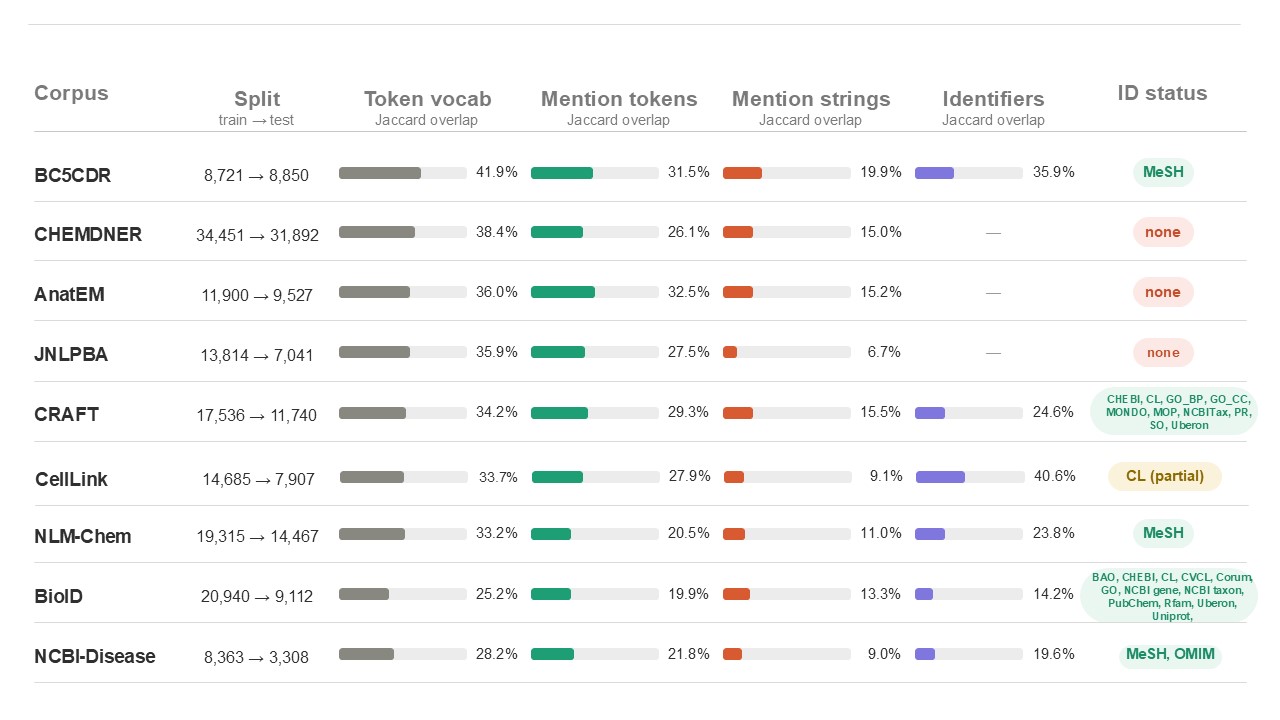}
  \caption{Train-test overlap across nine biomedical corpora. All values are Jaccard similarity (\%) between training and test splits. Token vocab: unique vocabulary tokens shared between splits. Mention tokens: unique tokens within entity mention spans. Mention strings: exact entity surface forms shared between splits. Identifiers: concept identifiers shared between splits.}
  \label{fig:overlap}
\end{figure*}

Figure~\ref{fig:overlap} reports train-test Jaccard overlap at four levels: token vocabulary, mention tokens, exact mention strings, and concept identifiers. 
These levels distinguish increasingly task-specific forms of reuse.
Across all corpora, token vocabulary overlap is higher than mention-token overlap, which is, in turn, higher than exact mention string overlap.

The slope of this drop varies and is itself informative.
JNLPBA falls from 35.9\% token overlap to 6.6\% mention-string overlap, indicating that its test split contains relatively little exact entity-form reuse.
CellLink presents a different structural pattern: its mention-token overlap (27.9\%) is much higher than its exact mention-string overlap (9.1\%), while identifier overlap is higher still (40.6\%).
This profile suggests compositional novelty, where new cell-population names are constructed from familiar component tokens and mapped to familiar concepts.
Examples include ``CD4+ T cells'' (CL:0000624), ``resting CD4 memory T cells'' (CL:0000897), and ``resting NK cells'' (CL:0000623), which combine recognizable modifiers with cell-type heads.

Identifier overlap adds a concept-level view unavailable from strings alone.
Among normalized corpora, identifier overlap ranges from 14.2\% in BioID to 40.6\% in CellLink---a range wider than the corresponding mention string overlaps, and with a different ordering.
CellLink's high identifier overlap paired with low string overlap means it tests the ability to map novel text to relatively familiar concepts.
Conversely, NCBI-Disease's low identifier overlap requires systems to generalize to genuinely novel conceptual territory.

\subsection{Metadata Composition}

Temporal coverage differs sharply:
CellLink is most recent, spanning 2019-2025;
CHEMDNER is nearly a single-year snapshot (97.7\% from 2013);
CRAFT spans 2001-2007; NCBI-Disease is predominantly pre-2000 (90.8\%); and BC5CDR is the broadest, spanning 1968-2016.

The journal distribution shows similar variation in domain.
BC5CDR is by far the most venue-diverse: 703 journals, with the top five journals accounting for only 5.0\% of the corpus.
BioID represents the opposite extreme, with 18 journals total and 83.2\% of corpus documents from the top five journals.

The article-topic mappings in Table~\ref{tab:topics} demonstrate substantial differences in the slice of biomedical literature these corpora represent.
NCBI-Disease is strongest in genetics/genomics, CHEMDNER and NLM-Chem emphasize chemistry/materials, BioID is dominated by molecular biology/biochemistry, and CellLink is enriched for general biology and cell/developmental biology.

\begin{table*}[t]
\centering
\small
\setlength{\tabcolsep}{2.3pt}
\begin{tabular}{lrrrrrrrrr}
\toprule
\textbf{Topic} & \textbf{AnatEM} & \textbf{BC5CDR} & \textbf{BioID} & \textbf{CHEM} & \textbf{CRAFT} & \textbf{CellLink} & \textbf{JNLPBA} & \textbf{NCBI} & \textbf{NLM} \\
\midrule
Molecular bio. / biochemistry & \textbf{16\%} & 6\% & \textbf{62\%} & 15\% & 21\% & 12\% & \textbf{36\%} & 17\% & 18\% \\
Chemistry / materials science & 8\% & 15\% & --- & \textbf{29\%} & 4\% & 4\% & 11\% & 5\% & \textbf{29\%} \\
Genetics/genomics & 5\% & --- & 8\% & 3\% & 18\% & 8\% & 10\% & \textbf{28\%} & 4\% \\
General bio. / anatomy / physiology & 14\% & 14\% & 6\% & 18\% & \textbf{23\%} & \textbf{22\%} & 11\% & 11\% & 14\% \\
Demographic characteristics & 7\% & \textbf{18\%} & --- & 4\% & 4\% & 5\% & 3\% & 16\% & 6\% \\
Cell \& developmental biology & 7\% & 1\% & 10\% & 5\% & 9\% & 17\% & 12\% & 4\% & 6\% \\
Clinical specialties by organ system & 8\% & 11\% & --- & 3\% & 7\% & 9\% & 2\% & 5\% & 4\% \\
General / internal medicine & 6\% & 6\% & 7\% & 2\% & 1\% & 3\% & 0\% & 1\% & 2\% \\
Pharmacology & 2\% & 6\% & --- & 6\% & --- & --- & 1\% & --- & 2\% \\
Other & 27\% & 23\% & 7\% & 15\% & 13\% & 20\% & 14\% & 13\% & 15\% \\
\bottomrule
\end{tabular}
\vspace{4pt}
\par\noindent\footnotesize
Topics are assigned from article MeSH terms where available; unresolved article terms fall back to NLM Catalog MeSH journal topics and configured journal-name anchors. ``Other'' aggregates all categories not shown. The largest displayed non-Other entry in each column is shown in bold.
\caption{Broad research topic composition of each corpus. Values represent approximate percentage of corpus articles from each topic area. CHEM = CHEMDNER; NCBI = NCBI-Disease; NLM = NLM-Chem.}
\label{tab:topics}
\end{table*}

\subsection{Terminology-based coverage}
\begin{figure*}[t]
  \centering
  \includegraphics[width=\linewidth]{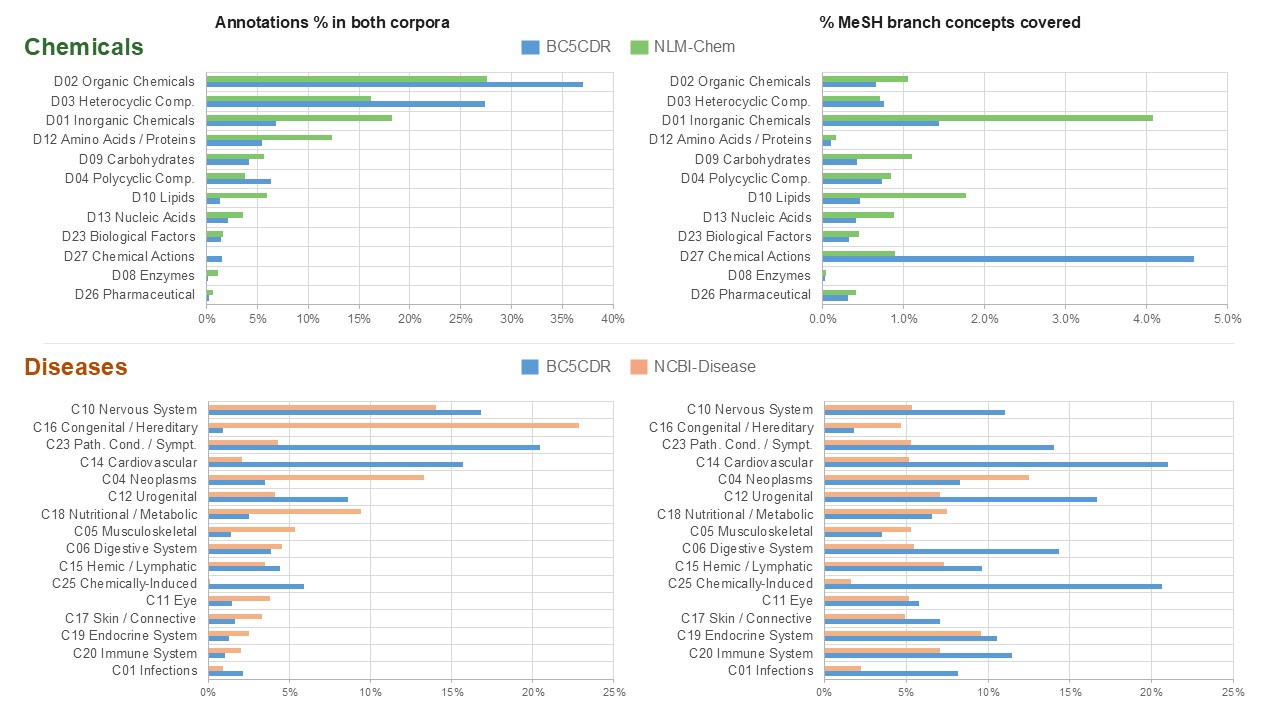}
  \caption{
  Terminology-based coverage analysis. Left panels: distribution normalized within each corpus's own annotations. Right panels: distribution normalized within the MeSH tree vocabulary. Top panels: distribution across MeSH chemical branches (D-branches) for BC5CDR and NLM-Chem. Bottom panels: distribution across MeSH disease branches (C-branches) for BC5CDR and NCBI-Disease.
}
  \label{fig:ontology}
\end{figure*}

Figure~\ref{fig:ontology} demonstrates that corpora sharing an entity label may still exercise different concept regions of the concept space. The left panels normalize branch counts within each corpus, showing what share of the corpus's own annotations falls into each high-level terminology branch. The right panels normalize the same branch counts by the size of the corresponding terminology branch, highlighting how the corpus represents different areas of the terminology.
Figure~\ref{fig:ontology} focuses on MeSH diseases and chemicals; analogous CL-based statistics for cell-type corpora are available in the dashboard.

For diseases, NCBI-Disease is dominated by Congenital and Hereditary Diseases (C16, 23\%) and Neoplasms (C04, 14\%), consistent with its genetics origin.
BC5CDR instead peaks at Pathological Conditions (C23, 20\%) and Cardiovascular Diseases (C14, 17\%), consistent with pharmacovigilance as a clinically cross-cutting focus rather than a narrow specialty focus.
Chemically-Induced Disorders account for 5.8\% of BC5CDR disease annotations but only 0.05\% of NCBI-Disease.
Thus, while both corpora evaluate “disease” recognition, they measure fundamentally distinct capabilities.

For chemicals, BC5CDR is concentrated in Organic Chemicals (D02, 37\%) and Heterocyclic Compounds (D03, 28\%), the categories containing most small-molecule drugs.
NLM-Chem is more broadly distributed, with higher representation of Inorganic Chemicals (D01, 18\%) and Amino Acids/Proteins (D12, 12\%).

\section{Discussion}

The central implication of this study is that benchmark results for biomedical NER and EL cannot be interpreted independently of the corpora that produced them. A benchmark corpus is not simply a labeled sample of text: it is a measurement instrument whose structure determines which system capabilities are exercised and limits how far evaluation conclusions can reasonably transfer. 
Because benchmark utility is task- and domain-dependent, resources that share an entity type label can still function as fundamentally different instruments, evaluating different capabilities, imposing different generalization demands, and representing different regions of biomedical literature and concept space.
No single statistic captures this multidimensionality. Instead, benchmark interpretation depends on the interaction of several diagnostic signals, and the nine corpora examined here illustrate how these signals combine in practice.

\textbf{Annotation density and label distribution.}
Annotation density determines how concentrated evaluation signal is within each sampled unit. Dense full-text corpora such as CRAFT and NLM-Chem yield many linked decisions from relatively few articles, which can improve evaluation sensitivity and expose systems to full-text sectional variation. 
At the same time, raw density is partly confounded with document length and annotation scope. A large number of annotations from one article may include repeated mentions of the same entities, and therefore may not provide the same independent evidence as the same number of annotations sampled from many distinct passages or documents. Sparse corpora based on abstracts, passages, or figure captions require broader sampling to achieve stable estimates, and they provide more independent contexts per annotation. Each design serves different evaluation purposes: full-text corpora emphasize document-level realism and repeated-use behavior, whereas passage- or abstract-based corpora can emphasize breadth of contexts under fixed annotation budgets.

\textbf{Lexical and conceptual variation.}
Variation determines whether a corpus tests recognition of repeated forms, alternative expressions for known concepts, or generalization to genuinely new ones. CellLink's variation of 3.74 surface forms per label-link pair reflects the compositional naming conventions of single-cell biology: novel cell-population names are assembled from familiar modifiers and cell-type heads, so systems must generalize across surface forms while mapping to relatively familiar concepts. BC5CDR's lower variation is consistent with the pharmacovigilance literature, where drug and disease names tend to be standardized.

\textbf{Train-test overlap.}
Overlap analysis offers a particularly direct diagnostic for interpreting unexpectedly strong or weak results, consistent with prior work on leakage and memorization ~\citep{Tutubalina2020,liang2023}. 
The steepness of the token-level overlap drop varies and is in itself informative. 
CellLink illustrates one diagnostic profile: high identifier overlap (40.6\%) combined with much lower mention-string overlap (9.1\%) means the benchmark emphasizes compositional surface-form generalization rather than novel concept recognition.
NCBI-Disease presents the opposite profile, with low identifier overlap (19.6\%) that places genuine demands on concept-level generalization. JNLPBA's sharp drop from 35.9\% token overlap to 6.7\% mention-string overlap, combined with the absence of normalization, 
may partially recontextualize its relatively low performance ceiling
[e.g.,~\citep{Huang2020}]: systems cannot rely heavily on exact entity-form reuse.

\textbf{Metadata composition.}
Temporal distributions, journal diversity, and topic profiles describe the slice of literature represented by the corpus. 
These properties define the domain over which benchmark claims can reasonably apply and determine whether conclusions transfer to corpora drawn from different time periods, venues, or subfields. CHEMDNER's near-total concentration in 2013 chemistry literature and BioID's extreme journal concentration (83.2\% from five journals) are structural features that may limit the generalizability of results. BC5CDR's breadth across 703 journals and five decades makes performance claims more domain-general.
These metadata properties are rarely reported yet have a bearing on whether performance on a benchmark supports claims about a target deployment setting.

\textbf{
Terminology coverage.
}The contrast between NCBI-Disease and BC5CDR illustrates why terminology-aware analysis matters even when two corpora nominally share a task label. Both annotate disease entities, yet NCBI-Disease is dominated by Congenital and Hereditary Diseases (C16) and Neoplasms (C04), reflecting its genetics origin, while BC5CDR peaks at Pathological Conditions (C23) and Cardiovascular Diseases (C14), reflecting its pharmacovigilance focus. Chemically-Induced Disorders (C25) account for 5.8\% of BC5CDR disease annotations but only 0.05\% of NCBI-Disease. High performance on one therefore supports only limited claims about transfer to the other, despite the shared entity label. The same logic applies to chemical corpora: BC5CDR's concentration in Organic Chemicals and Heterocyclic Compounds reflects a small-molecule drug emphasis, while NLM-Chem's broader distribution across Inorganic Chemicals, Amino Acids, and Lipids reflects the wider biochemical scope of full-text molecular biology literature. 
Differences in terminology coverage may reflect different practical capabilities even when their reported task names are identical.

\textbf{
From diagnostics to corpus decisions.} 
Corpus diagnostics are informative not only retrospectively but before a benchmark is finalized. Before releasing a split, developers can compute mention-string and identifier overlap to detect leakage. While extending a corpus, they can sample documents to fill temporal, journal, topic, or terminology-branch gaps. When selecting a benchmark, researchers can choose corpora whose density, overlap profile, and terminology coverage match the intended deployment claim. When combining corpora, they can verify whether a new source adds concept regions not already represented
or mainly duplicates familiar strings and identifiers.
The goal, then, is to tie benchmark interpretation to corpus characteristics and intended use, rather than assign a single quality score.

\textbf{Limitations.}
The framework characterizes corpus structure but does not directly predict downstream system rankings or benchmark saturation. Measured ambiguity conflates true polysemy, underspecified guidelines, and annotation errors; the present framework does not distinguish among these sources. Surface-form variation measures observed diversity rather than the full range of synonymy present in the literature. Terminology analyses depend on the version of the reference ontology used. Finally, topic mappings provide practical, lightweight domain diagnostics rather than definitive subject classifications.

\section{Conclusion}

We presented a corpus-centric framework for diagnosing the benchmark utility of biomedical corpora with entity annotations. The framework organizes standardized statistics over annotations, linked identifiers, corpus splits, document metadata, and terminology mappings into five diagnostic families: (1) scale, density and label distribution, (2) lexical and conceptual structure, (3) train-test overlap, (4) metadata composition, and (5) terminology coverage. Applied to nine biomedical NER and EL corpora, these diagnostics show that corpora for the same apparent task can measure substantially different capabilities.

The main implication is that these statistics provide a practical basis for more complete reporting of biomedical NER and EL benchmarks.
Corpus reports should describe annotation density, normalization support, lexical and identifier variation, train-test overlap at multiple abstraction levels, temporal and journal composition, and terminology coverage where applicable. Without these measurements, benchmark results are difficult to interpret.
Standardized corpus diagnostics would make evaluation claims more interpretable, clarify what a benchmark measures, and support transparent, reproducible, transfer-aware evaluation.

\section*{Acknowledgments}

This research was supported by the Intramural Research Program of the National Institutes of Health (NIH). The NIH author contributions are considered Works of the United States Government. The findings and conclusions presented are those of the authors and do not necessarily reflect the views of the NIH or the U.S. Department of Health and Human Services.

\bibliography{custom}

@inproceedings{Arighi2017,
  author    = "Arighi, Cecilia and Lynette Hirschman and Thomas Lemberger and Samuel Bayer and Robin Liechti and Donald Comeau and Cathy Wu",
  title     = "{Bio-ID track overview}",
  booktitle = "BioCreative VI Challenge Evaluation Workshop",
  year      = "2017",
  address   = "Bethesda, MD",
  pages     = "14-19"
}

@article{aroyo2015,
author = {Aroyo, Lora and Welty, Chris},
title = {Truth Is a Lie: Crowd Truth and the Seven Myths of Human Annotation},
year = {2015},
issue_date = {Spring 2015},
publisher = {John Wiley \& Sons, Inc.},
address = {USA},
volume = {36},
number = {1},
issn = {0738-4602},
url = {https://doi.org/10.1609/aimag.v36i1.2564},
doi = {10.1609/aimag.v36i1.2564},
journal = {AI Magazine},
month = mar,
pages = {15–24},
numpages = {10}
}

@article{artstein2008,
  title={Survey Article: Inter-Coder Agreement for Computational Linguistics},
  author={Artstein, Ron and Poesio, Massimo},
  journal={Computational Linguistics},
  volume={34},
  number={4},
  pages={555--596},
  year={2008},
  doi="10.1162/coli.07-034-R2",
  url="https://aclanthology.org/J08-4004/"
}

@article{Bada2012,
  author  = "Bada, Michael and Miriam Eckert and Donald Evans and Kristin Garcia and Krista Shipley and Dmitry Sitnikov and William A Baumgartner Jr and K Bretonnel Cohen and Karin Verspoor and Judith A Blake and Lawrence E. Hunter",
  title   = "{Concept annotation in the CRAFT corpus.}",
  journal = "BMC Bioinformatics",
  year    = "2012",
  volume  = "13",
  number  = "",
  pages   = "161",
  doi     = "10.1186/1471-2105-13-161"
}

@inproceedings{Collier2004,
  author    = "Collier, Nigel and Tomoko Ohta and Yoshimasa Tsuruoka and Yuka Tateisi and Jin-Dong Kim",
  title     = "Introduction to the Bio-entity Recognition Task at {JNLPBA}",
  booktitle = "Proceedings of the International Joint Workshop on Natural Language Processing in Biomedicine and its Applications (NLPBA/BioNLP)",
  year      = "2004",
  pages     = "73-78",
  URL       = "https://aclanthology.org/W04-1213/",
  publisher = "COLING",
  Address   = "Geneva, Switzerland"
}

@article{Comeau2013,
  author    = "Donald C. Comeau and Rezarta Islamaj Doğan and Paolo Ciccarese and Kevin Bretonnel Cohen and Martin Krallinger and Florian Leitner and Zhiyong Lu and Yifan Peng and Fabio Rinaldi and Manabu Torii and Alfonso Valencia and Karin Verspoor and Thomas C. Wiegers and Cathy H. Wu and W. John Wilbur",
  title     = "{BioC: a minimalist approach to interoperability for biomedical text processing.}",
  journal   = "Database (Oxford)",
  pages     = "bat064",
  year      = "2013",
  doi       = "10.1093/database/bat064",
  volume    = "2013"
}

@article{Comeau2019,
    author = {Comeau, Donald C and Wei, Chih-Hsuan and Islamaj Doğan, Rezarta and Lu, Zhiyong},
    title = {{PMC} text mining subset in {BioC}: about three million full-text articles and growing},
    journal = {Bioinformatics},
    volume = {35},
    number = {18},
    pages = {3533-3535},
    year = {2019},
    month = {09},
    issn = {1367-4803},
    doi = {10.1093/bioinformatics/btz070},
    url = {https://doi.org/10.1093/bioinformatics/btz070},
    eprint = {https://academic.oup.com/bioinformatics/article-pdf/35/18/3533/48975610/bioinformatics_35_18_3533.pdf},
}

@article{Dogan2014,
  author  = "Doğan, Rezarta Islamaj and Robert Leaman and Zhiyong Lu",
  title   = "{NCBI disease corpus: A resource for disease name recognition and concept normalization.}",
  journal = "Journal of Biomedical Informatics",
  year    = "2014",
  volume  = "47",
  number  = "",
  pages   = "1–10",
  doi     = "10.1016/j.jbi.2013.12.006", 
  url     = {https://www.sciencedirect.com/science/article/pii/S1532046413001974}
}

@inproceedings{fries2022,
  title={{BigBio}: A framework for data-centric biomedical natural language processing},
  author={Fries, Jason and Weber, Leon and Seelam, Natasha and Altay, Gabriel and Datta, Debajyoti and Garda, Samuele and Kang, Sunny and Su, Rosaline and Kusa, Wojciech and Cahyawijaya, Samuel and others},
  booktitle={Advances in Neural Information Processing Systems},
  volume={35},
  pages={25792--25806},
  year={2022},
  url={https://proceedings.neurips.cc/paper_files/paper/2022/hash/a583d2197eafc4afdd41f5b8765555c5-Abstract-Datasets_and_Benchmarks.html}
}

@article{Gu2021,
  author    = {Gu, Yu and Tinn, Robert and Cheng, Hao and Lucas, Michael and Usuyama, Naoto and Liu, Xiaodong and Naumann, Tristan and Gao, Jianfeng and Poon, Hoifung},
  title     = "{Domain-Specific Language Model Pretraining for Biomedical Natural Language Processing.}",
  journal = "ACM Transactions on Computing for Healthcare ({HEALTH})",
  volume  = "3",
  issue   = "1",
  pages   = "1--23",
  year    = "2021",
  doi     = {10.1145/3458754}
}

@inproceedings{Gururangan2018,
  author    = {Gururangan, Suchin and Swayamdipta, Swabha and Levy, Omer and Schwartz, Roy and Bowman, Samuel R. and Smith, Noah A.},
  title     = "{Annotation Artifacts in Natural Language Inference Data, New Orleans, Louisiana, Association for Computational Linguistics.}",
  booktitle = {Proceedings of the 2018 Conference of the North American Chapter of the Association for Computational Linguistics: Human Language Technologies, Volume 2 (Short Papers)},
  pages = {107--112},
  address = {New Orleans, Louisiana},
  publisher = {Association for Computational Linguistics},
  year = {2018},
  doi = {10.18653/v1/N18-2017}
}

@article{Herrero-Zazo2013,
  author  = {Herrero-Zazo, María and Segura-Bedmar, Isabel and Martínez, Paloma and Declerck, Thierry},
  title   = "{The {DDI} corpus: an annotated corpus with pharmacological substances and drug-drug interactions}",
  journal = {Journal of Biomedical Informatics},
  volume = {46},
  number = {5},
  pages = {914--920},
  year = {2013},
  doi = {10.1016/j.jbi.2013.07.011}
}

@article{Hovy2010,
  title={Towards a `science' of corpus annotation: a new methodological challenge for corpus linguistics},
  author={Hovy, Eduard and Lavid, Julia},
  journal={International Journal of Translation},
  volume={22},
  number={1},
  pages={13--36},
  year={2010}
}

@article{hripcsak2005,
  title={Agreement, the f-measure, and reliability in information retrieval},
  author={Hripcsak, George and Rothschild, Adam S},
  journal={Journal of the American Medical Informatics Association},
  volume={12},
  number={3},
  pages={296--298},
  year={2005},
  doi = {10.1197/jamia.M1733}
}

@article{Huang2020,
  title={Biomedical named entity recognition and linking datasets: survey and our recent development},
  author={Huang, Ming-Siang and Lai, Po-Ting and Lin, Pei-Yen and You, Yu-Ting and Tsai, Richard Tzong-Han and Hsu, Wen-Lian},
  journal={Briefings in Bioinformatics},
  volume={21},
  number={6},
  pages={2219--2238},
  year={2020},
  publisher={Oxford University Press},
  doi={10.1093/bib/bbaa054}
}

@article{Islamaj2021chem,
  author = {Islamaj, Rezarta and Leaman, Robert and Kim, Sun and Kwon, Dongseop and Wei, Chih-Hsuan and Comeau, Donald C. and Peng, Yifan and Cissel, David and Coss, Cathleen and Fisher, Carol and Guzman, Rob and Kochar, Preeti Gokal and Koppel, Stella and Trinh, Dorothy and Sekiya, Keiko and Ward, Janice and Whitman, Deborah and Schmidt, Susan and Lu, Zhiyong},
  title   = "{NLM-Chem, a new resource for chemical entity recognition in PubMed full text literature.}",
  journal = "Scientific Data",
  year    = "2021",
  volume  = "8",
  number  = "1",
  pages   = "91",
  doi = {10.1038/s41597-021-00875-1}
}

@article{Islamaj2021gene,
  author  = "Islamaj and R. and C. H. Wei and D. Cissel and N. Miliaras and O. Printseva and O. Rodionov and K. Sekiya and J. Ward and Z. Lu",
  title   = "{NLM-Gene, a richly annotated gold standard dataset for gene entities that addresses ambiguity and multi-species gene recognition.}",
  journal = "Journal of Biomedical Informatics",
  year    = "2021",
  volume  = "118",
  number  = "",
  pages   = "103779",
  doi     = {10.1016/j.jbi.2021.103779}
}

@article{Islamaj2024b,
  author    = "Islamaj and R. and C. H. Wei and P. T. Lai and L. Luo and C. Coss and P. Gokal Kochar and N. Miliaras and O. Rodionov and K. Sekiya and D. Trinh and D. Whitman and Z. Lu",
  title     = "{The biomedical relationship corpus of the BioRED track at the BioCreative VIII challenge and workshop.}",
  journal = "Database (Oxford)",
  volume  = {2024},
  year      = "2024",
  pages      = "baae099",
  doi       = {10.1093/database/baae071}
}

@article{Kim2003,
  author    = "Kim, Jin-Dong and Ohta, Tomoko and Tateisi, Yuka and Tsujii, Jun'ichi",
  title     = "{GENIA corpus--a semantically annotated corpus for bio-textmining}",
  journal = "Bioinformatics",
  volume    = {19},
  pages    = {i180-i182},
  issue    = {suppl_1},
  doi    = {10.1093/bioinformatics/btg1023},
  year      = "2003"
}

@article{Krallinger2015,
  author = {Krallinger, Martin and Rabal, Obdulia and Leitner, Florian and Vazquez, Miguel and Salgado, David and Lu, Zhiyong and Leaman, Robert and Lu, Yanan and Ji, Donghong and Lowe, Daniel M and Sayle, Roger A and Batista-Navarro, Riza Theresa and Rak, Rafal and Huber, Torsten and Rocktäschel, Tim and Matos, Sérgio and Campos, David and Tang, Buzhou and Hua Xu and Tsendsuren Munkhdalai and Keun Ho Ryu and SV Ramanan and Senthil Nathan and Slavko Žitnik and Marko Bajec and Lutz Weber and Matthias Irmer and Saber A Akhondi and Jan A Kors and Shuo Xu and Xin An and Utpal Kumar Sikdar and Asif Ekbal and Masaharu Yoshioka and Thaer M Dieb and Miji Choi and Karin Verspoor and Madian Khabsa and C Lee Giles and Hongfang Liu and Komandur Elayavilli Ravikumar and Andre Lamurias and Francisco M Couto and Hong-Jie Dai and Richard Tzong-Han Tsai and Caglar Ata and Tolga Can and Anabel Usié and Rui Alves and Isabel Segura-Bedmar and Paloma Martínez and Julen Oyarzabal and Alfonso Valencia},
  title     = "{The CHEMDNER corpus of chemicals and drugs and its annotation principles}",
  journal = "Journal of Cheminformatics",
  volume      = "7",
  issue      = "Suppl 1",
  pages      = "S2",
  year      = "2015",
  doi       = {10.1186/1758-2946-7-S1-S2}
}

@article{Li2016,
  author = {Li, Jiao and Sun, Yueping and Johnson, Robin J. and Sciaky, Daniela and Wei, Chih-Hsuan and Leaman, Robert and Davis, Allan Peter and Mattingly, Carolyn J. and Wiegers, Thomas C. and Lu, Zhiyong},
  title = {{BioCreative V CDR} task corpus: a resource for chemical disease relation extraction},
  journal = {Database},
  volume = {2016},
  pages = {baw068},
  year = {2016},
  doi = {10.1093/database/baw068}
}

@article{Liang2023,
  title     = {Holistic Evaluation of Language Models},
  author    = {Percy Liang and Rishi Bommasani and Tony Lee and Dimitris Tsipras and Dilara Soylu and Michihiro Yasunaga and Yian Zhang and Deepak Narayanan and Yuhuai Wu and Ananya Kumar and Benjamin Newman and Binhang Yuan and Bobby Yan and Ce Zhang and Christian Cosgrove and Christopher D Manning and Christopher Re and Diana Acosta-Navas and Drew A. Hudson and Eric Zelikman and Esin Durmus and Faisal Ladhak and Frieda Rong and Hongyu Ren and Huaxiu Yao and Jue WANG and Keshav Santhanam and Laurel Orr and Lucia Zheng and Mert Yuksekgonul and Mirac Suzgun and Nathan Kim and Neel Guha and Niladri S. Chatterji and Omar Khattab and Peter Henderson and Qian Huang and Ryan Andrew Chi and Sang Michael Xie and Shibani Santurkar and Surya Ganguli and Tatsunori Hashimoto and Thomas Icard and Tianyi Zhang and Vishrav Chaudhary and William Wang and Xuechen Li and Yifan Mai and Yuhui Zhang and Yuta Koreeda},
  journal   = {Transactions on Machine Learning Research},
  year      = {2023},
  url       = {https://openreview.net/forum?id=iO4LZibEqW}
}

@article{Lipscomb2000,
  title={Medical Subject Headings ({MeSH})},
  author={Lipscomb, Carolyn E},
  journal={Bulletin of the Medical Library Association},
  volume={88},
  number={3},
  pages={265--266},
  year={2000}
}

@article{Lu2011,
  author = {Lu, Zhiyong and Kao, Hung-Yu and Wei, Chih-Hsuan and Huang, Minlie and Liu, Jingchen and Kuo, Cheng-Ju and Hsu, Chun-Nan and Tsai, Richard Tzong-Han and Dai, Hong-Jie and Okazaki, Naoaki and Cho, Hye-Cheol and Gerner, Martin and Solt, Illes and Agarwal, Shashank and Liu, Feifan and Vishnyakova, Dina and Ruch, Patrick and Romacker, Martin and Rinaldi, Fabio and Sanmitra Bhattacharya and Padmini Srinivasan and Hongfang Liu and Manabu Torii and Sergio Matos and David Campos and Karin Verspoor and Kevin M Livingston and W John Wilbur},
  title     = "{The gene normalization task in BioCreative III}",
  journal   = "BMC Bioinformatics",
  volume    = "12",
  issue     = "Suppl 8",
  pages     = "S2",
  year      = "2011",
  doi       = "10.1186/1471-2105-12-S8-S2"
}

@article{Malik2026,
  title={{ChEBI}: re-engineered for a sustainable future},
  author={Malik, Adnan and Arsalan, Muhammad and Moreno, Carlos and Mosquera, Juan and F{\'e}lix, Eloy and Kizil{\"o}ren, Tevfik and Muthukrishnan, Venkatesh and Zdrazil, Barbara and Leach, Andrew R and O’Boyle, Noel M},
  journal={Nucleic Acids Research},
  volume={54},
  issue={D1},
  pages={D1768--D1778},
  year={2026},
  publisher={Oxford University Press},
  doi={10.1093/nar/gkaf1271}
}

@article{miranda2023,
  author    = "Miranda-Escalada and A. and F. Mehryary and J. Luoma and D. Estrada-Zavala and L. Gasco and S. Pyysalo and A. Valencia and M. Krallinger",
  title     = "{Overview of DrugProt task at BioCreative VII: data and methods for large-scale text mining and knowledge graph generation of heterogenous chemical-protein relations.}",
  journal = "Database (Oxford)",
  volume={2023},
  pages={baad080},
  year      = "2023",
  doi ={10.1093/database/baad080}
}

@article{Morgan2008,
  author = {Morgan, Alexander A. and Lu, Zhiyong and Wang, Xinglong and Cohen, Aaron M. and Fluck, Juliane and Ruch, Patrick and Divoli, Anna and Fundel, Katrin and Leaman, Robert and Hakenberg, Jörg and Sun, Chenjie and Liu, Heng-hui and Torres, Rafael and Krauthammer, Michael and Lau, William W. and Liu, Hongfang and Hsu, Chun-Nan and Schuemie, Martijn and Cohen, K. Bretonnel and Hirschman, Lynette},
  title = {Overview of {BioCreative II} gene normalization},
  journal = {Genome Biology},
  volume = {9},
  number = {Suppl 2},
  pages = {S3},
  year = {2008},
  doi = {10.1186/gb-2008-9-s2-s3}
}

@inproceedings{Ogren2006,
    title = "{K}nowtator: A Prot{\'e}g{\'e} plug-in for annotated corpus construction",
    author = "Ogren, Philip V.",
    editor = "Rudnicky, Alex  and
      Dowding, John  and
      Milic-Frayling, Natasa",
    booktitle = "Proceedings of the Human Language Technology Conference of the {NAACL}, Companion Volume: Demonstrations",
    month = jun,
    year = "2006",
    address = "New York City, USA",
    publisher = "Association for Computational Linguistics",
    url = "https://aclanthology.org/N06-4006/",
    pages = "273--275"
}

@inproceedings{Peng2019,
  author = {Peng, Yifan and Yan, Shankai and Lu, Zhiyong},
  title = {Transfer learning in biomedical natural language processing: An evaluation of {BERT} and {ELMo} on ten benchmarking datasets},
  booktitle = {Proceedings of the 18th BioNLP Workshop and Shared Task},
  pages = {58--65},
  address = {Florence, Italy},
  publisher = {Association for Computational Linguistics},
  year = {2019},
  doi = {10.18653/v1/W19-5006}
}

@article{Pyysalo2014,
  author = {Pyysalo, Sampo and Ananiadou, Sophia},
  title = {Anatomical entity mention recognition at literature scale},
  journal = {Bioinformatics},
  volume = {30},
  number = {6},
  pages = {868--875},
  year = {2014},
  doi = {10.1093/bioinformatics/btt580}
}

@article{Rotenberg2026,
  author = "Rotenberg and N. H. and R. Leaman and R. Islamaj and H. Kuivaniemi and G. Tromp and B. Fluharty and S. Richardson and C. Eastwood and M. Diller and B. Xu and A. V. Pankajam and D. Osumi-Sutherland and Z. Lu and R. H. Scheuermann",
  title = "{Cell phenotypes in the biomedical literature: a systematic analysis and text mining corpus.}",
  journal="bioRxiv",
  year = "2026",
  doi = {10.64898/2026.02.11.705457},
}

@inproceedings{Stenetorp2012,
  title={brat: a Web-based Tool for NLP-Assisted Text Annotation},
  author={Stenetorp, Pontus and Pyysalo, Sampo and Topi{\'c}, Goran and Ohta, Tomoko and Ananiadou, Sophia and Tsujii, Jun’ichi},
  booktitle={Proceedings of the Demonstrations at the 13th Conference of the European Chapter of the Association for Computational Linguistics},
  pages={102--107},
  address = {Avignon, France},
  publisher = {Association for Computational Linguistics},
  url = {https://aclanthology.org/E12-2021/},
  year={2012}
}

@article{Tan2026,
  title = {The Cell Ontology in the age of single-cell omics},
  author = {Tan, Shawn Zheng Kai and Puig-Barbe, Aleix and Goutte-Gattat, Damien and Eastwood, Caroline and Aevermann, Brian and Avola, Alida and Balhoff, James P. and Bayindir, Ismail Ugur and Belfiore, Jasmine and Caron, Anita Reane and David S. Fischer and Nancy George and Benjamin M. Gyori and Melissa A. Haendel and Charles Tapley Hoyt and Huseyin Kir and Tiago Lubiana and Nicolas Matentzoglu and James A. Overton and Beverly Peng and Bjoern Peters and Ellen M. Quardokus and Patrick L. Ray and Paola Roncaglia and Andrea D. Rivera and Ray Stefancsik and Wei Kheng Teh and Sabrina Toro and Nicole Vasilevsky and Chuan Xu and Yun Zhang and Richard H. Scheuermann and Christopher J. Mungall and Alexander D. Diehl and David Osumi-Sutherland},
  journal = {Scientific Data},
  year = {2026},
  doi = {10.1038/s41597-026-07173-8}
}

@inproceedings{Tutubalina2020,
    title = "Fair Evaluation in Concept Normalization: a Large-scale Comparative Analysis for {BERT}-based Models",
    author = "Tutubalina, Elena  and
      Kadurin, Artur  and
      Miftahutdinov, Zulfat",
    editor = "Scott, Donia  and
      Bel, Nuria  and
      Zong, Chengqing",
    booktitle = "Proceedings of the 28th International Conference on Computational Linguistics",
    month = dec,
    year = "2020",
    address = "Barcelona, Spain (Online)",
    publisher = "International Committee on Computational Linguistics",
    url = "https://aclanthology.org/2020.coling-main.588/",
    doi = "10.18653/v1/2020.coling-main.588",
    pages = "6710--6716"
}

@article{Uma2021,
author = {Uma, Alexandra and Fornaciari, Tommaso and Hovy, Dirk and Paun, Silviu and Plank, Barbara and Poesio, Massimo},
year = {2021},
month = {12},
pages = {1385-1470},
title = {Learning from Disagreement: A Survey},
volume = {72},
journal = {Journal of Artificial Intelligence Research},
doi = {10.1613/jair.1.12752}
}

@article{Vasilevsky2026,
  title={Mondo: integrating disease terminology across communities},
  author={Nicole A Vasilevsky and Sabrina Toro and Nicolas Matentzoglu and Joseph E Flack and Kathleen R Mullen and Harshad Hegde and Sarah Gehrke and Patricia L Whetzel and Yousif Shwetar and Nomi L Harris and Mee S Ngu and Gioconda L Alyea and Megan S Kane and Paola Roncaglia and Eric Sid and Courtney L Thaxton and Valerie Wood and Roshini S Abraham and Maria Isabel Achatz and Pamela Ajuyah and Joanna S Amberger and Lawrence Babb and Jasmine Baker and James P Balhoff and Jonathan S Berg and Amol Bhalla and Xavier Bofill-De Ros and Ian R Braun and Eleanor C Broeren and Blake K Byer and Alicia B Byrne and Tiffany J Callahan and Leigh C Carmody and Lauren E Chan and Amanda R Clause and Julie S Cohen and Marcello DeLuca and Natalie T Deuitch and May Flowers and Jamie Fraser and Toyofumi Fujiwara and Vanessa Gitau and Jennifer L Goldstein and Dylan Gration and Tudor Groza and Benjamin M Gyori and William Hankey and Jason A Hilton and Daniel S Himmelstein and Stephanie S Hong and Charles T Hoyt and Robert Huether and Eric Hurwitz and Julius O B Jacobsen and Atsuo Kikuchi and Sebastian Köhler and Daniel R Korn and David Lagorce and Bryan J Laraway and Jane Y Li and Adriana J Malheiro and James McLaughlin and Birgit H M Meldal and Shruthi Mohan and Sierra A T Moxon and Monica C Munoz-Torres and Tristan H Nelson and Frank W Nicholas and David Ochoa and Daniel Olson and Tudor I Oprea and Tomiko T Oskotsky and David Osumi-Sutherland and Kelley Paris and Helen E Parkinson and Zoë M Pendlington and Xiao P Peng and Amy Pizzino and Sharon E Plon and Bradford C Powell and Julie C Ratliff and Heidi L Rehm and Lyubov Remennik and Erin R Riggs and Sean Roberts and Peter N Robinson and Justyne E Ross and Kevin Schaper and Brian M Schilder and Johanna L Schmidt and Elliott W Sharp and Morgan N Similuk and Damian Smedley and Tam P Sneddon and Rachel Sparks and Ray Stefancsik and Gregory S Stupp and Shilpa Sundar and Terue Takatsuki and Imke Tammen and Kezang C Tshering and Deepak R Unni and Eloise Valasek and Adeline Vanderver and Alex H Wagner and Ryan F Webb and Danielle Welter and Doron Yaya-Stupp and Andreas Zankl and Xingmin Aaron Zhang and Julie A McMurry and Christopher G Chute and Ada Hamosh and Christopher J Mungall and Melissa A Haendel ClinGen DICER1 and miRNA-Processing Gene Variant Curation Expert Panel and ClinGen Hereditary Gene Curation Expert Panel and ClinGen Motile Ciliopathy Gene Curation Expert Panel and ClinGen Myeloid Malignancy Variant Curation Expert Panel and ClinGen TP53 Variant Curation Expert Panel and ClinGen X-Linked Inherited Retinal Disease Variant Curation Expert Panel},
  journal={Genetics},
  volume={232},
  number={4},
  pages={iyaf215},
  year={2026},
  publisher={Oxford University Press US},
  doi={10.1093/genetics/iyaf215}
}

@inproceedings{Wang2019,
  author = {Wang, Alex and Pruksachatkun, Yada and Nangia, Nikita and Singh, Amanpreet and Michael, Julian and Hill, Felix and Levy, Omer and Bowman, Samuel},
  booktitle = {Advances in Neural Information Processing Systems},
  title = {{SuperGLUE}: A Stickier Benchmark for General-Purpose Language Understanding Systems},
  volume      = "32",
  year      = "2019",
  url = {https://proceedings.neurips.cc/paper/2019/hash/4496bf24afe7fab6f046bf4923da8de6-Abstract.html}
}

@inproceedings{Wang2018,
    title = "{GLUE}: A Multi-Task Benchmark and Analysis Platform for Natural Language Understanding",
    author = "Wang, Alex  and
      Singh, Amanpreet  and
      Michael, Julian  and
      Hill, Felix  and
      Levy, Omer  and
      Bowman, Samuel R.",
    editor = "Linzen, Tal  and
      Chrupa{\l}a, Grzegorz  and
      Alishahi, Afra",
    booktitle = "Proceedings of the 2018 {EMNLP} Workshop {B}lackbox{NLP}: Analyzing and Interpreting Neural Networks for {NLP}",
    month = nov,
    year = "2018",
    address = "Brussels, Belgium",
    publisher = "Association for Computational Linguistics",
    url = "https://aclanthology.org/W18-5446/",
    doi = "10.18653/v1/W18-5446",
    pages = "353--355"
}

@article{Wei2013,
  author  = "Wei, Chih-Hsuan and Harris, Bethany R. and Kao, Hung-Yu and Lu, Zhiyong",
  title   = "{tmVar: a text mining approach for extracting sequence variants in biomedical literature.}",
  journal = "Bioinformatics",
  year    = "2013",
  volume  = "29",
  number  = "11",
  pages   = "1433--1439",
  doi = {10.1093/bioinformatics/btt156}
}

@article{Wei2016,
  author    = {Wei, Chih-Hsuan and Peng, Yifan and Leaman, Robert and Davis, Allan Peter and Mattingly, Carolyn J. and Li, Jiao and Wiegers, Thomas C. and Lu, Zhiyong},
  title     = "{Assessing the state of the art in biomedical relation extraction: overview of the BioCreative V chemical-disease relation ({CDR}) task.}",
  journal = {Database (Oxford)},
  volume = {2016},
  pages = {baw032},
  year = {2016},
  doi = {10.1093/database/baw032}
}

@article{wei2024,
    author = {Wei, Chih-Hsuan and Allot, Alexis and Lai, Po-Ting and Leaman, Robert and Tian, Shubo and Luo, Ling and Jin, Qiao and Wang, Zhizheng and Chen, Qingyu and Lu, Zhiyong},
    title = {{PubTator} 3.0: an {AI}-powered literature resource for unlocking biomedical knowledge},
    journal = {Nucleic Acids Research},
    volume = {52},
    number = {W1},
    pages = {W540-W546},
    year = {2024},
    month = {07},
    issn = {0305-1048},
    doi = {10.1093/nar/gkae235},
    url = {https://doi.org/10.1093/nar/gkae235},
    eprint = {https://academic.oup.com/nar/article-pdf/52/W1/W540/58436124/gkae235.pdf},
}

\end{document}